\documentclass[runningheads]{llncs}
\usepackage{microtype}
\usepackage{graphicx}
\usepackage{comment}
\usepackage{amsmath, amssymb} \usepackage{color}
\usepackage{pifont}

\usepackage{epsfig}
\usepackage{wrapfig}
\usepackage{subcaption}
\usepackage{dblfloatfix}
\usepackage{times}
\usepackage{cite}
\usepackage[font=small,labelfont=bf]{caption}
\usepackage{cancel}
\usepackage{paralist}
\usepackage{chngcntr}

\usepackage{multirow}
\usepackage{rotating}
\usepackage{booktabs}
\usepackage[table]{xcolor}
\usepackage{tabularx}

\usepackage[pagebackref=true,breaklinks=true,letterpaper=true,colorlinks,bookmarks=false]{hyperref}

\begin{document}
\newcommand{\aman}[1]{{\color{red}[AS: #1]}}
\newcommand{\vedanuj}[1]{{\color{blue}[VG: #1]}}
\newcommand{\devi}[1]{{\color{purple}[DP: #1]}}
\newcommand{\add}[1]{{\color{orange}#1}}
\newcommand{\dpc}[1]{{\color{purple}#1}}
\pagestyle{headings}
\mainmatter
\def\ECCVSubNumber{4360}  
\title{Are we pretraining it right? \\ Digging deeper into visio-linguistic pretraining} 

\titlerunning{Are we pretraining it right? Digging deeper into visio-linguistic pretraining}
\author{Amanpreet Singh\thanks{Equal contribution}\inst{1} \and
Vedanuj Goswami$^\star$\inst{1} \and
Devi Parikh\inst{1,2}}
\authorrunning{A. Singh et al.}
\institute{Facebook AI Research\and
Georgia Institute of Technology \\
\email{ \{asg,vedanuj\}@fb.com} \email{parikh@gatech.edu}}
\maketitle

\begin{abstract}

Numerous recent works have proposed pretraining generic visio-linguistic representations and then finetuning them for downstream vision and language tasks. While architecture and objective function design choices have received attention, the choice of pretraining datasets has received little attention. In this work, we question some of the default choices made in literature. For instance, we systematically study how varying similarity between the pretraining dataset domain (textual and visual) and the downstream domain affects performance. Surprisingly, we show that automatically generated data in a domain closer to the downstream task (e.g., VQA v2) is a better choice for pretraining than ``natural'' data but of a slightly different domain (e.g., Conceptual Captions). On the other hand, some seemingly reasonable choices of pretraining datasets were found to be entirely ineffective for some downstream tasks. This suggests that despite the numerous recent efforts, vision \& language pretraining does not quite work ``out of the box'' yet. Overall, as a by-product of our study, we find that simple design choices in pretraining can help us achieve close to state-of-art results on downstream tasks without any architectural changes.

\end{abstract}

\section{Introduction}
Vision and language tasks such as image captioning and visual question answering have witnessed tremendous progress in recent years, with CNN and RNN fusion-based models rapidly improving the state-of-the-art on benchmarks. This breakthrough in supervised models led to the inundation of labelled datasets but the field has entered a stage of saturation where the performance increases log-linearly with the size of labelled data \cite{vqa2challenge2019}. Moreover, domains for which collecting large quantities of labelled data is hard have been deprived of significant gains. \cite{singh2019towards, gurari2018vizwiz, shah2019kvqa, biten2019scene}.

In Natural Language Understanding (NLU), the recent success of self-supervised learning has significantly changed the research landscape by achieving state-of-the-art results on various low-resource benchmarks \cite{wang2019superglue, wang2018glue, devlin2018bert, radford2019language}. 
Naturally, this shift has also influenced visio-linguistic architectures and training significantly \cite{lu2019vilbert, li2019visualbert, tan2019lxmert, chen2019uniter, lu201912, alberti2019fusion, li2019unicoder, su2019vl, zhou2019unified}. Contrary to language model training where a word corpus is directly used, in visio-linguistic self-supervised pretraining, a combination of images and text is used. A pretraining proxy task with a self-supervised objective is used to train the model to predict some hidden (masked) part of the input whether it is an image feature or a word from the text. Large image captioning datasets have been a go-to choice as a pretraining dataset as they provide detailed descriptions of an image which can then be used to learn task-agnostic and generic language grounding in images \cite{chen2015microsoft, sharma2018conceptual, ordonez2011im2text}. Finally, the model is fine-tuned end-to-end on a downstream task by replacing the head of the pre-trained network with task-specific heads.

\begin{figure}[t]
\centering
\includegraphics[width=\columnwidth]{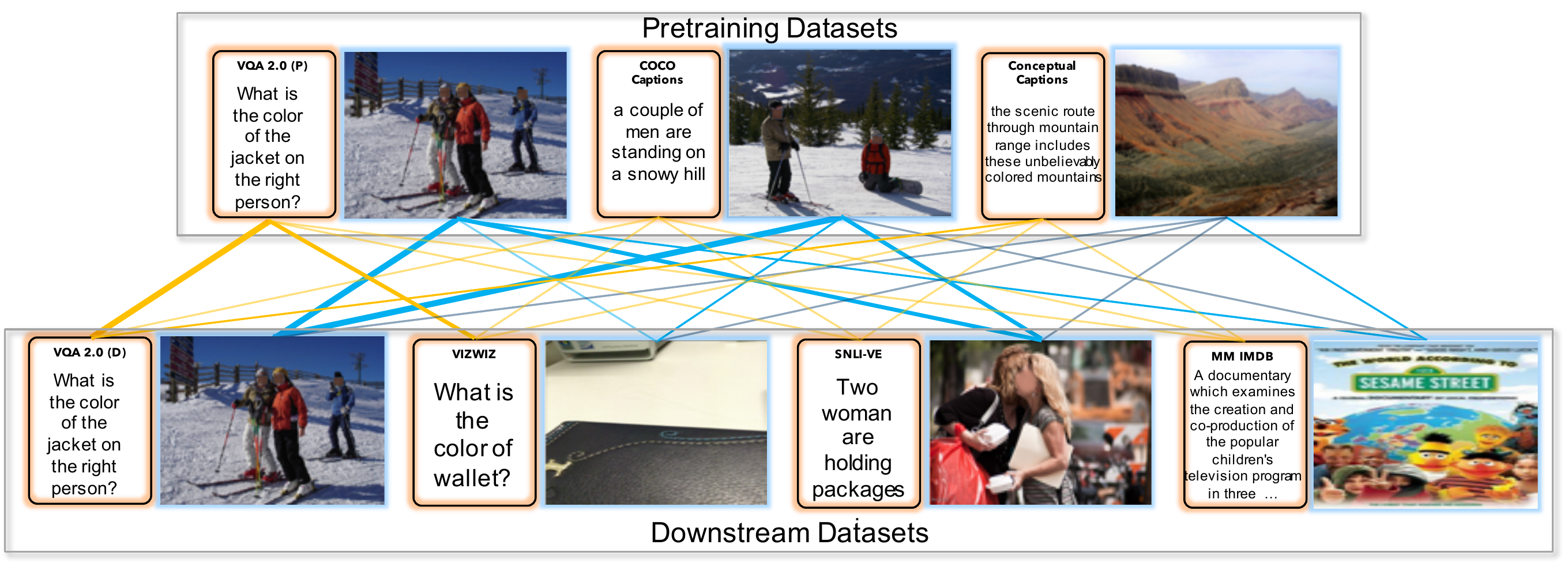}
\caption{Illustration of domain similarity between pretraining datasets and downstream tasks that are considered in this study. Node represents textual and visual domains of the datasets and edges the domain similarities (blue for visual and orange for textual). Thicker edges show higher  domain similarity. In this study, we investigate which factors are important for choosing the pretraining dataset and how we can increase the similarity between pretraining and downstream domains for better downstream performance.}
  \vspace{-8 pt}
\label{fig:intro}
\end{figure}

The design choices made while pretraining these architectures have been based on intuition for the most part. The finer nuances and details of the pretrain-finetune regime have not been carefully investigated. For instance, a lot of recent works use Conceptual Captions\cite{sharma2018conceptual} as the pretraining dataset due to its large size. But perhaps COCO Captions\cite{cococaption}, due to its less noisy nature, would be a better fit? Should the domain of the downstream task be considered when deciding which pretraining dataset will be the most effective? Is automatically generated data in a domain closer to the downstream task a better choice for pretraining than ``natural'' data but of a more different domain?

In this work, as a step towards answering these questions, first we carefully choose a set of pretraining datasets and downstream tasks. The pretraining datasets are chosen with varying degrees of similarities in textual and visual domains to the downstream tasks. This allows us to understand the effects of varying the domain of pretraining datasets on downstream tasks. For pretraining we choose VQA 2.0 \cite{goyal2017making}, COCO Captions\cite{cococaption} and Conceptual Captions (CC) \cite{sharma2018conceptual}. For downstream tasks we choose VQA\cite{VQA}, Vizwiz\cite{gurari2018vizwiz}, SNLI-VE\cite{xie2018visual} and MM-IMDB\cite{arevalo2017gated}. Fig.\ref{fig:intro} represents these datasets as graph nodes with thickness of the edge connecting them specifying how similar two datasets are in either visual or textual domains. For instance, the VQA pretraining dataset has a similar textual domain (natural language questions) as Vizwiz but different visual domain (images taken by blind users instead of well-composed images on the web). On the other hand VQA and COCO have same visual domains (in fact, same set of images) but different textual domains (questions vs. captions).

We empirically show how the domain of the pretraining dataset affects the downstream task performance. Further, we do a deeper dive to show what makes pretraining representations transfer well to downstream tasks and how low-resource tasks can benefit from this effectively. As a further step, we try to improve accuracy of downstream tasks by pushing the domains of the pretraining dataset and downstream tasks closer. We achieve this by generating a synthetic dataset which is closer in domain to the downstream task. Interestingly, our synthetic dataset achieves better performance on the downstream task compared to a more ``natural'' commonly used dataset that is a worse match in domain to the downstream task. This is an important result because it has the potential to help overcome the limitation of scalability when pretraining visio-linguistic representations which relies on paired labelled data.

To summarize, we make the following primary contributions:
\begin{compactitem}
    \item We do a deeper dive into the intricacies of pretraining visio-linguistic architectures and show through extensive empirical analysis the importance of choosing the right setup for pretraining.
    \item Leveraging the above findings we show how some simple design choices in pretraining can help us achieve close to state-of-art results on downstream tasks without any architectural changes.
    \item We generate a synthetic dataset that is closer to the domain of a downstream task, and show that pretraining on this synthetic dataset results in higher accuracies on the downstream task than a commonly used ``natural'' dataset.     This opens up significant potential to overcome the limitation of scaling up visio-linguistic pretraining which relies largely on paired data.
\end{compactitem}

\section{Study Setup}
\label{sec:setup}
In this section, we describe the setup of our study and motivate our key choices. We start by introducing our choice of pretraining datasets in Section~\ref{subsec:pretraining_datasets}. In Section~\ref{subsec:downstream_tasks}, we describe the downstream datasets on which we evaluate along with their metrics. In Section~\ref{subsec:models}, we describe the transformer-based architectures that we have chosen for our study and then we discuss the pretraining objectives that we use for training these models in Section~\ref{subsec:objectives}. Finally, we conclude this section by providing details of our training and experimentation setup in Section~\ref{subsec:setup}.

\subsection{Pretraining Datasets}
\label{subsec:pretraining_datasets}
We consider three different pretraining datasets with an aim to cover different aspects that can matter during pretraining, including but not limited to scale/size, quality of images/annotations and visual and textual domain distributions.

\textbf{COCO Captions} \cite{cococaption}. Common Objects in Context \cite{lin2014microsoft} are natural images collected from Flickr which contain common scenes from daily life. COCO was introduced with bounding boxes, segmentation masks and keypoints for 80 common categories to advance state-of-the-art in object detection and segmentation. COCO Captions \cite{cococaption} was later collected to complement progress in multimodal AI. COCO Captions contains 200k labelled images and each image has a set of five captions which provides a total of 1M image-caption pairs.

\textbf{VQA 2.0} \cite{goyal2017making}. The Visual Question Answering~\cite{VQA} task involves understanding and reasoning about an image to answer a question. VQA 1.0~\cite{VQA} was collected on COCO images~\cite{lin2014microsoft}. VQA 2.0 \cite{goyal2017making} was later introduced to balance the language-biases created by questions which could be answered without even looking at the image (e.g., ``What color is the banana?'' Yellow). Complementary images were provided for each question such that both images were similar, but had different answers to the same question. VQA 2.0  contains 1.1M questions on 200k images from COCO. Each question has 10 human-annotated answers. 
\textbf{Conceptual Captions} \cite{sharma2018conceptual}. Conceptual Captions (CC) is a collection of 3.3M image-caption pairs scraped from the web by pairing images with their associated alternate text. CC provides large diversity and scale in visual content. On the other hand, in spite of the effort to clean the dataset, the automatic collection process (understandably) results in captions that are of poorer quality than those in COCO. In our experiments, we use a subset of 3.1M image-caption pairs from CC which are currently available for download.

COCO and VQA 2.0 share the same image source (COCO) but have different textual domains (captions vs questions). Comparing performance on a downstream task after pretraining on these two datasets allows us to understand the effect of varying textual domains. Similarly, COCO and CC share the textual domain\footnote{Though both are captions, Conceptual Captions in a lot of examples don't feel natural while COCO Captions are always human annotated} (captions) but come from very different image sources (common objects in daily life vs wikipedia style images); this helps us quantify the impact of the visual domain. Finally, VQA 2.0 and CC have both different visual and textual domains. 
\subsection{Downstream Datasets}
\label{subsec:downstream_tasks}
To fully evaluate the impact of domain of pretraining datasets on transferability to downstream tasks, we would ideally want to explore downstream datasets that cover all four combinations of \{textual,visual\} $\times$ \{match,mismatch\} to the pretraining datasets. Of course, it is difficult to talk about match vs. mismatch in domains in a binary sense, but we describe below our attempt to span this space as we select the four downstream datasets we experiment with.

\textbf{VQA 2.0} \cite{goyal2017making}. In the VQA task, for a given image and question pair $(I, Q)$, an approach has to predict an answer $A$, usually from a fixed vocabulary. We use the VQA 2.0 dataset for this task. We described this dataset earlier as a pretraining dataset, but it can also be used as a downstream VQA task.  The evaluation is performed using the VQA Accuracy metric\footnote{More details of the metric can be found here: \href{https://visualqa.org/evaluation.html}{https://visualqa.org/evaluation.html}}.
Hereafter, we refer to the downstream VQA 2.0 task as VQA-D and the pretraining VQA 2.0 dataset as VQA-P to avoid confusion.

\textbf{VizWiz} \cite{gurari2018vizwiz}. The VizWiz dataset contains 32K images from blind users collected using the VizWiz app \cite{bigham2010vizwiz}. We use the question-answering task from VizWiz as a downstream task in which blind users ask questions on these images to address some of their daily needs. Each of the 32K question has 10 sighted human-annotated answers and the VQA accuracy is used as the evaluation metric. Notably, 54\% of the questions are \textit{unanswerable} because the image may be irrelevant to the question or too blurry. The real-world nature of the data makes this task challenging.

\textbf{SNLI-VE} \cite{xie2018visual}. The SNLI-VE(\textbf{SNLI} \textbf{V}isual \textbf{E}ntailment) dataset is generated based on SNLI \cite{bowman2015large} and Flickr30k \cite{plummer2015flickr30k} datasets. Flickr30k comes from the same image source as COCO (i.e., Flickr). Given an image and a natural language statement, the visual entailment task involves classifying whether the statement is true (entailment), false (contradiction) or neutral \textit{w.r.t.} to the image. The dataset contains 550K image/statement pairs and evaluation is done using classification accuracy. 

\textbf{MM-IMDB} \cite{arevalo2017gated}. The MM-IMDB(\textbf{M}ulti \textbf{M}odal \textbf{IMDB}) dataset consists of 26K movie plot outlines and movie posters. The task involves assigning genres to each movie from a list of 23 genres. This is a multilabel prediction problem, i.e., one movie can have multiple genres and we use micro-F1 and macro-F1 as evaluation metrics following \cite{arevalo2017gated}.

\begin{figure}[t]
\centering
\includegraphics[width=\columnwidth]{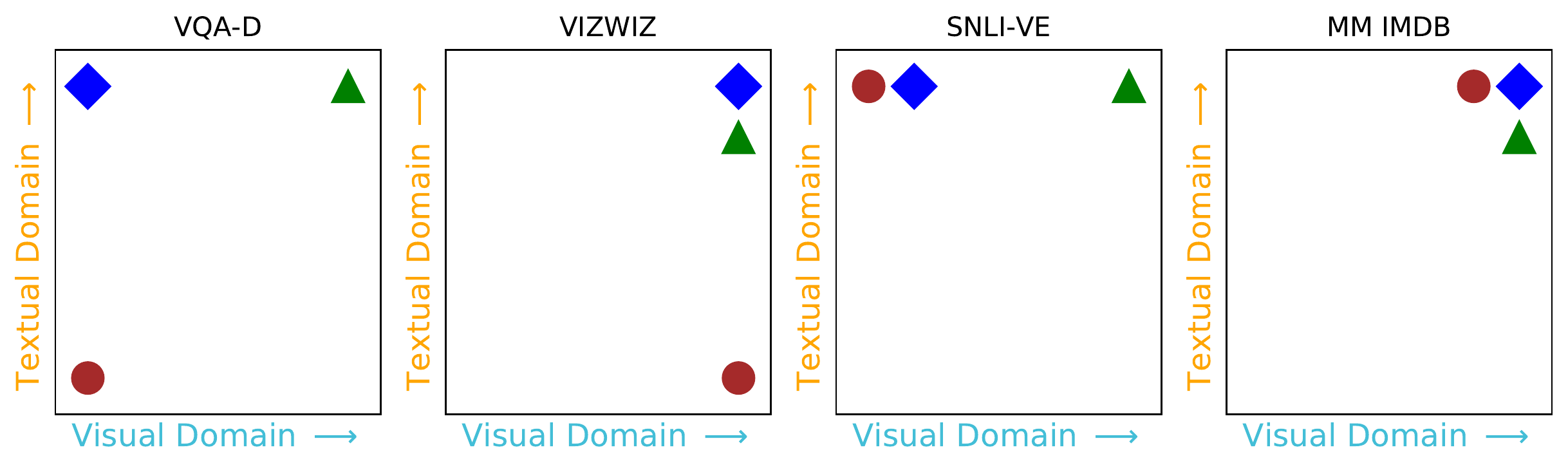}
\includegraphics[width=\columnwidth]{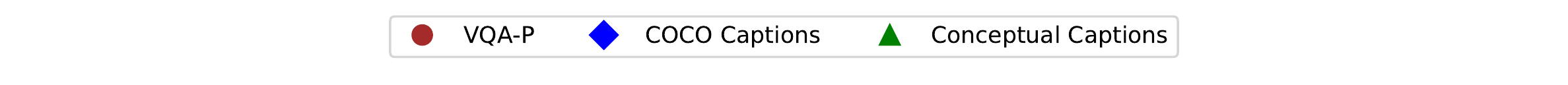}
\caption{A visualization of how different pretraining datasets (symbols in the figure) relate to downstream tasks (headings) in the 2D space formed by  visual and textual domain similarity}
  \vspace{-8 pt}
\label{fig:domain_similarity}
\end{figure}

Fig.~\ref{fig:domain_similarity} shows how the domains of the different pretraining and downstream datasets are related. VQA-D and SNLI-VE share the visual domain with COCO while VizWiz and VQA-D share the textual domain with VQA-P. Since, VQA-P and VQA-D are the same dataset, they share both the textual and visual domains. SNLI-VE does not share the textual domain with any of the pretraining datasets while VizWiz does not share the visual domain with any. MM-IMDB shares neither the textual nor the visual domain with any of the pretraining datasets; results on it will demonstrate the transferability of pretrained representations when both textual and visual domains do not match. This perhaps most closely mimics real world scenarios. Finally, CC, one of the pretraining datasets, does not share either of the textual or visual domains with any of the downstream datasets. This is primarily because to the best of our knowledge, none of the existing vision and language benchmarks match CC's domains. Moreover, this provides us an opportunity to study transferability of pretrained representations across mismatched domains.

MM-IMDB and VizWiz will also help us in testing the transferability of pretrained representations to downstream datasets that are small, analogous to low-resource scenarios in the real world.  As far as we know, the recent work in pretraining visio-linguistic representations evaluates only on large datasets such as VQA-P \cite{goyal2017making} and SNLI-VE \cite{xie2018visual} by fine-tuning the entire  model end-to-end. This masks the true generalizability of the learnt visio-linguistic representations.

We acknowledge that as with conclusions drawn from any empirical evaluation, our findings are specific to the pretraining and downstream datasets we chose. But, we believe our choices cover a wide span of scenarios which makes our findings valuable.

\subsection{Models}
\label{subsec:models}
From the recent plethora of visio-linguistic pretraining approaches\cite{lu2019vilbert, li2019visualbert, li2019unicoder, alberti2019fusion, su2019vl, chen2019uniter, kiela2019supervised}, two major categories of architectures have emerged. Single-stream architectures like VisualBERT \cite{li2019visualbert} project and convert both visual and textual embeddings into a single embedding space before passing them through the transformer layers. On the other hand, dual-stream architectures like ViLBERT \cite{lu2019vilbert} pass the embeddings separately through different transformers and merge them at the end. We experiment with both VisualBERT and ViLBERT to cover both classes of architectures and different model capacities. The base VisualBERT model (ignoring task-specific heads) has 110M parameters and ViLBERT has 250M. 
\textbf{VisualBERT} \cite{li2019visualbert} is a single stream BERT model \cite{devlin2018bert} with multiple transformer blocks \cite{vaswani2017attention}. The image regions's embeddings concatenated with textual embeddings are the input to the model in a similar fashion as BERT but twice as wide. The image embeddings are computed by adding image region embeddings, image positional embeddings and a specific embedding which distinguishes it from the text embeddings. VisualBERT's input looks like \texttt{[CLS]}, \texttt{l$_1$}, \dots, \texttt{[MASK]}, \dots, \texttt{l$_M$}, \texttt{[SEP]}, \texttt{v$_1$}, \dots, \texttt{v$_N$}. Similar to BERT \texttt{l$_i$} represents a textual input token, \texttt{[MASK]} and \texttt{[SEP]} represent the masked input and separator tokens used in self-supervised pretraining. \texttt{v$_i$} represents an object embedding extracted from the image. This joint input is passed through the transformer blocks and the final representation corresponding to \texttt{[CLS]} token is used in downstream tasks.

\textbf{ViLBERT} \cite{lu2019vilbert} consists of two parallel BERT transformer streams connected by co-attention transformer (\texttt{TRM}) block layers. One stream of transformer blocks are for the visual input and the other for the linguistic input. For image input $I$ represented as a set of region features $v_i$ and textual input $l_i$, the model generates representations $h_{vi}$ and $h_{li}$. The co-attention \texttt{TRM} blocks are added for specific layers between the visual and textual \texttt{TRM} blocks. In comparison to VisualBERT, input to ViLBERT has two separate inputs of the form ``\texttt{[CLS]}, \texttt{t$_1$}, \dots, \texttt{[MASK]}, \texttt{t$_M$}'' and ``\texttt{v$_1$}, \texttt{v$_2$}, \dots, \texttt{[MASK]}, \dots, \texttt{v$_N$}''. Note that in ViLBERT's training, even image objects are masked and predicted in the pretraining objective. Similar to VisualBERT, \texttt{[CLS]} token's final representation is used in any downstream tasks.

\subsection{Pretraining Objectives}
\label{subsec:objectives}
Following \cite{li2019visualbert} and \cite{lu2019vilbert}, we use two types of pretraining objectives. 
\textbf{Masked Language Modeling (MLM).} Recall that image regions are $\mathbf{v} = \{v_1, ..., v_N\}$, and the input words are $\mathbf{l} = \{ l_1, ..., l_M\}$. The objective is to reconstruct $\mathbf{l}$ from a corrupted version $\mathbf{\hat{l}}$ where some words $\mathbf{l}_\mathbf{m}$ are masked i.e. replaced with a \texttt{[MASK]} token randomly with probability $p$. Let $\theta$ be the trainable parameters. We minimize the negative log-likelihood:
\begin{equation}
    \mathcal{L}_{\text{MLM}}(\theta) = -E_{(\mathbf{l}, \mathbf{v})\sim D} \log P_{\theta}(\mathbf{l} |\mathbf{\hat{l}}, \mathbf{v}).
\end{equation}

\textbf{Masked Multimodal Modeling (MMM).} Similar to MLM, let us denote the image regions as $\mathbf{v} = \{v_1, ..., v_N\}$, the input texts as $\mathbf{l} = \{ l_1, ..., l_M\}$. In MMM, the objective is to reconstruct either $\mathbf{l}$ and/or $\mathbf{v}$ from corrupted  versions $\mathbf{\hat{v}}$ and $\mathbf{\hat{l}}$ where some words $\mathbf{l}_\mathbf{m}$ or image regions  $\mathbf{v}_\mathbf{n}$ are masked. Image regions are \textit{masked} with probability $p_v$ by setting their features to zeros. Masked text words are replaced by a \texttt{[MASK]} token randomly with probability, $p_l$. We minimize the negative log-likelihood:
\begin{equation}
    \mathcal{L}_{\text{MMM}}(\theta) = E_{(\mathbf{l}, \mathbf{v})\sim D} \log P_{\theta}(\mathbf{l} |\mathbf{\hat{l}}, \mathbf{\hat{v}}) + E_{(\mathbf{l}, \mathbf{v})\sim D} f_{\theta}(\mathbf{v} | \mathbf{\hat{v}}, \mathbf{\hat{l}}).
\end{equation}
where $\theta$ are the trainable parameters and $f_{\theta}$ is a region class prediction network. MMM can be considered as a combination of MLM and MIM (masked image modeling) objectives.

We use MLM for training VisualBERT, and MMM for training ViLBERT following the original papers \cite{li2019visualbert,lu2019vilbert}. We drop the image-sentence alignment objective from \cite{lu2019vilbert} as it can't be used with VQA-P where same question can be correct for multiple images. We also drop the next sentence prediction objective from \cite{li2019visualbert} as it can't be used with CC which only has one image. For VQA-P, we use question answer pairs as the text to pretrain. Specifically, the input looks like ``\texttt{[CLS]}, \texttt{l$_1$}, \dots, \texttt{l$_M$}, \texttt{[SEP]}, \texttt{A}'' where \texttt{A} is randomly sampled from 10 human-annotated answers. The input is then masked at random with probability \textit{p}.
\subsection{Experimental Setup}
\label{subsec:setup}
We use the Pythia framework \cite{singh2018pythia,singh2019towards} for our experiments which is based on PyTorch\footnote{Pythia is available at \href{https://github.com/facebookresearch/pythia}{https://github.com/facebookresearch/pythia}}. For VisualBERT and ViLBERT, we take the original implementations and incorporate them inside Pythia ensuring no implementation differences. We train our models in a distributed fashion on 4 nodes each containing 8 NVIDIA V100 GPUs. We conducted extensive hyperparameter search and used the best configuration wherever possible for each experiment whether pretraining or finetuning. For ViLBERT, we use the default setting of 6 and 12 \texttt{TRM} blocks for the visual and linguistic streams respectively as used in \cite{lu2019vilbert}. For VisualBERT, we follow \cite{li2019visualbert} and use 12 \texttt{TRM} blocks.

Both VisualBERT and ViLBERT models are first initialized from pretrained BERT weights provided by the HuggingFace Transformers library \cite{Wolf2019HuggingFacesTS}. Specifically, we initialize 12 \texttt{TRM} layers of VisualBERT and 12 language \texttt{TRM} layers of ViLBERT. We extract 2048$\mathcal{D}$ region based image features from \texttt{fc6} layer of a ResNeXT-152 based Faster-RCNN model \cite{xie2017resnext,ren2015faster} trained on Visual Genome \cite{krishna2017visual} with the attribute prediction loss following \cite{anderson2018bottom}. We do not use grid based features as used in previous works \cite{singh2019towards} due to incompatibility with MMM and for simplicity. For fine-tuning via back-propagation \cite{lecun1990handwritten} on downstream tasks, we use binary cross entropy loss to support multi-label predictions for VizWiz, VQA-D and MM-IMDB while we use cross entropy loss for three-way classification on SNLI-VE. We evaluate every 6000 updates and report the model with the best evaluation metric on the validation set. We use the AdamW optimizer \cite{kingma2014adam, loshchilov2017decoupled} with a cosine warmup and cosine decay learning rate scheduler. For AdamW, we set the value of $\epsilon$ to 1e-8 and ($\beta_1$, $\beta_2$) to (0.9, 0.999). Following \cite{ma2019adequacy}, we set warmup iterations to always be 2000. We use a learning rate of 5e-5, batch of size 1024 and set training update steps to 88k unless otherwise specified\footnote{More details on hyperparameters for each of the experiments are in supplementary}. We plan to release model weights from our experiments as well as the code. The value of \textit{p}, \textit{p$_v$} and \textit{p$_l$} are set to follow masking probabilities as in original BERT paper \cite{devlin2018bert}.

For a fair comparison between different pretraining datasets, we clip the number of samples present in VQA-P and CC to be the same as COCO. The chosen samples were randomly selected. CC is significantly larger than COCO. So, we also experiment with various sizes of CC ranging from 10\% of CC (which is the same size as COCO) to all of CC (CC-full). Hereafter, we refer to the smaller clipped CC dataset as CC-small.

\section{Life, the Universe and Pretraining}
\begin{table}[t]
\centering
  \caption{\textbf{Performance} on different downstream datasets when we \textbf{finetune} pretrained models. When downstream dataset's visual and textual domain match with the pretraining dataset, we observe maximum performance. CC-Small pretraining performs worst in most cases because its visual and textual domains are different from all of the downstream tasks. Pretraining doesn't work on MM-IMDB as its visual and textual domain don't match with any of our pretraining datasets.}
\smallskip
\setlength\tabcolsep{10 pt}
\resizebox{\columnwidth}{!}{
  \begin{tabular}{c | c | c c c c c }
  \toprule
      \multicolumn{1}{c}{} & \multicolumn{1}{c}{} & \multicolumn{5}{c}{Finetuned On}\\
    \cmidrule(r){3-7}
     Pretrained on & Model & VQA 2.0 (D) & Vizwiz & SNLI-VE & \multicolumn{2}{c}{MM IMDB} \\
    \cmidrule(lr){3-3}
    \cmidrule(lr){4-4}
    \cmidrule(lr){5-5}
    \cmidrule(lr){6-7}
     & & acc & acc & acc & macro F1 & micro F1 \\
      \midrule
    - & VisualBERT & 68.28 & 52.45 & 75.93 & \bf 60.02 & \bf 68.14 \\
    COCO & VisualBERT & \bf 69.90  & 53.19 & \bf 77.57 &  58.08 & 66.47 \\
    VQA 2.0 (P) & VisualBERT & 69.34 & \bf 53.44 & 77.20 & 57.79 & 66.50 \\
    CC-Small & VisualBERT & 68.58 & 52.59 & 77.44 & 58.19 & 67.04 \\
    \\      \midrule
     - & ViLBERT & 68.15 & 51.59 & 75.16  & \bf 58.48 & \bf 66.77 \\
    COCO & ViLBERT & 69.01  & \bf 52.84 & \bf 75.78 & 57.70 & 66.42 \\
    VQA 2.0 (P) & ViLBERT & \bf 69.05 & 52.77 & 75.39 & 57.72 & 65.63 \\
    CC-Small & ViLBERT & 68.42 & 52.04 & 69.40 & 58.20 & 66.70 \\
    \\  \bottomrule
  \end{tabular}}
  \label{tab:finetune}
\end{table}

In this section, we study different questions, one at a time, about methodology, generalizability/transferability, scalability, visual and textual domain impact and other often overlooked design choices for pretraining visio-linguistic representations. In each subsection below, we first motivate the question, detail the experiments we run to answer the question, present the results, and discuss the empirical trends and associated conclusions for that question. Finally, we show how all of these insights can be combined to achieve two near-SoTA and two SoTA results on downstream tasks. 
\subsection{How should one choose the pretraining dataset?}
\label{subsec:pretraining_importance}
In this subsection, we empirically evaluate and answer two questions (i) To what extent is the effectiveness of pretrained  visio-linguistic representations \textbf{agnostic to the visual and textual domains} of the downstream task? 
(ii) Is always using the \textbf{largest available dataset} for pretraining a good rule of thumb? We first pretrain both model architectures on each of our three pretraining datasets separately.  Table~\ref{tab:finetune} shows results of fine-tuning best pretrained models from all three pretraining datasets on each of the four downstream tasks. We also include results on direct training (without pretraining). 
\textbf{Visual and Textual Domain.} We observe that on downstream datasets where COCO and VQA-P match either or both visual and textual domains (VQA-D, VizWiz, SNLI-VE), they outperform CC-Small in results for both models. On VQA-D and VizWiz, both COCO and VQA-P perform competitively with each other. On SNLI-VE, where COCO matches the visual domain, it always outperforms both the other pretraining datasets. Interestingly, on MM-IMDb, direct training works better than pretraining on any of the datasets. These results strongly suggest that pretraining may not always work if there is a \textit{mismatch} between the domain of pretraining and downstream dataset \textit{i.e.} \textbf{using in-domain pretraining dataset is better} than using out-of-domain datasets.

\begin{figure}[ht]
    \centering
    \includegraphics[width=\columnwidth]{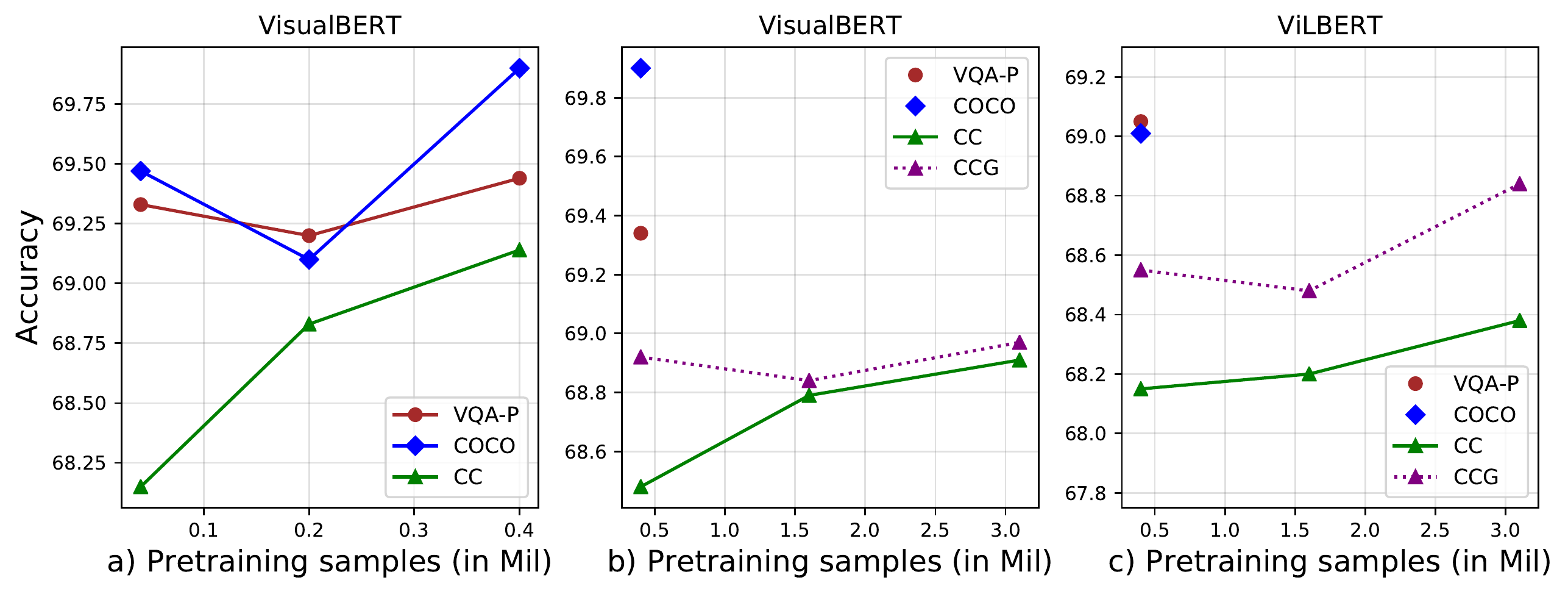}
    \caption{ \textbf{(a)} \textbf{Performance with differently sized subsets} of three pretraining datasets VQA-P, COCO and CC on VQA-D . COCO and VQA-P pretraining consistently outperform CC pretraining on all data sizes. \textbf{(b \& c)} Comparison of VQA-D accuracy (with finetuning) after pretraining on \textbf{CC Generated(CCG) vs VQA-P, COCO and CC} along with performance on differently sized subsets of CCG and CC using b) VisualBERT c) ViLBERT. CCG pretraining consistently outperforms CC even as we increase data but is always worse than COCO and VQA-P pretraining.}
      \vspace{-8 pt}
    \label{fig:cc_vs_cc_generated}
\end{figure}

\textbf{Size.} We study the effect of scaling up the pretraining data size for all the three pretraining datasets: VQA-P, COCO and CC-Small. We train on different subsets [0.04M, 0.2M, 0.4M] for each of the datasets. We keep all the other settings exactly same for a fair comparison. Fig.~\ref{fig:cc_vs_cc_generated}(a) shows the effect of increasing dataset size for the three pre-training datasets on downstream task VQA-D.
We observe an increasing trend and more pretraining data improves downstream task performance for all the three datasets. We also observe that for same amount of data, in-domain datasets (VQA-P, COCO) perform better than out-of-domain (CC) dataset consistently. Interestingly, for VQA-P and COCO increasing pre-training dataset size from 0.04M to 0.2M shows a diminishing trend but again improves when we move to 0.4M (full) size of the datasets.

We acknowledge that CC's primary feature is its size. To account for that, we pretrain both models for 30 epochs on different sized subsets of CC: 10\% (same as CC-small), 50\%, 100\% of the full set. 
We compare performance of fine-tuning these models on VQA-D in Fig.~\ref{fig:cc_vs_cc_generated}(b) and Fig.~\ref{fig:cc_vs_cc_generated}(c) using VisualBERT and ViLBERT respectively. We observe that COCO and VQA-P still outperform CC by a good margin even when its full scale is used. This suggests that dataset size is not the most important factor in visio-linguistic pretraining, even among existing options of datasets. With the right visual and textual domain match with the downstream task along with good quality, pretraining on even a \textit{smaller} dataset can easily outperform pretraining on a larger dataset.

\begin{table}[t]
\centering
\footnotesize
\caption{\textbf{Performance} on different downstream datasets when we \textbf{freeze the base} of the model. VQA-P pretraining consistently outperforms other pretraining datasets by a large margin suggesting VQA-P has more transferable representations. The results on MM-IMDB are inconsistent probably because both visual and textual domain mismatch with all pretraining datasets.}
\smallskip
\setlength\tabcolsep{10 pt}
\resizebox{\columnwidth}{!}{
  \begin{tabular}{c | c | c c c c c }
  \toprule
      \multicolumn{1}{c}{} & \multicolumn{1}{c}{} & \multicolumn{5}{c}{Finetuned (only head, base is frozen) On}\\
    \cmidrule(r){3-7}
     Pretrained on & Model & VQA 2.0 (D) & Vizwiz & SNLI-VE & \multicolumn{2}{c}{MM IMDB} \\
    \cmidrule(lr){3-3}
    \cmidrule(lr){4-4}
    \cmidrule(lr){5-5}
    \cmidrule(lr){6-7}
     & & acc & acc & acc & macro F1 & micro F1 \\
      \midrule
    - & VisualBERT & 43.47 & 44.32 & 34.73 & 2.99 & 30.86 \\
    COCO & VisualBERT & 48.64  & 44.38  & 54.13 & \bf 40.22 & \bf 55.01 \\
    VQA 2.0 (P) & VisualBERT & \textbf{60.25} & \textbf{47.23} & \textbf{56.85}  & 36.58 & 52.01 \\
    CC-Small & VisualBERT & 48.30 & 45.20  & 48.13  & 30.19 & 49.65\\
    \\      \midrule
     - & ViLBERT & 33.39 & 40.25 & 46.60 & 16.80 & 40.53 \\
    COCO & ViLBERT & 30.21 & 40.04 &  43.83 & 20.07 & 40.94 \\
    VQA 2.0 (P) & ViLBERT & \bf 52.97 & \bf 44.24 & \bf 49.92 & 20.94 & 41.69 \\
    CC-Small & ViLBERT & 26.88 & 39.43 & 40.97 & \bf 26.06 & \bf 46.17 \\
    \\  \bottomrule
  \end{tabular}}
  \vspace{-1mm}
  \label{tab:freeze}
\end{table}

\subsection{Does pretraining always help?}
\label{subsec:does_it_help}

\indent\indent\textbf{Low-resource downstream tasks.} In visio-linguistic pretraining, recent literature evaluates on large downstream datasets such as VQA-D \cite{goyal2017making} and SNLI-VE \cite{xie2018visual} by fine-tuning the entire model end-to-end. This masks the actual transferability and contribution of pretrained visio-linguistic representations. To study this,
we look at results on two small downstream datasets, VizWiz \cite{gurari2018vizwiz} and MM-IMDB \cite{arevalo2017gated} from Section~\ref{subsec:pretraining_importance}.
In Table~\ref{tab:finetune}, we observe that performance on VizWiz improves when we pretrain on VQA-P in which the text domain (questions) matches. Surprisingly, for MM-IMDB, which has different text and image domain compared to all of our pretraining datasets, we achieve best performance when no pretraining is used. This suggests that pretraining does not always help, and one should be mindful of characteristics of the downstream task.

\textbf{Transferability.} To better understand whether visio-linguistic pretraining learns something relevant for downstream tasks, we freeze the base of the model and only finetune the classifier head. This allows us to directly measure transferability to the downstream task without any task-specific finetuning. From the performance gap in Table~\ref{tab:freeze}, it is evident that compared to randomly initialized embeddings (row 1), pretraining does learn more transferable features. However, the actual transferability varies between different datasets. Surprisingly, we find that when pretrained with COCO and CC-Small and frozen (i.e., no finetuning) ViLBERT applied to all downstream tasks doesn't benefit from pretraining. We see the same in results on SNLI-VE as the downstream task when pretrained on CC-small even if the model is fine-tuned and not frozen (Table~\ref{tab:finetune}).

\subsection{Which pretrained representations are more transferable?}
\label{subsec:transferability}
From Table~\ref{tab:freeze}, we observe that a model pretrained on VQA-P learns better representations that can be transferred to downstream tasks like VQA-D, Vizwiz and SNLI-VE for both VisualBERT and ViLBERT models. We hypothesize that VQA-P is a very good dataset for learning transferable representation because it is more diverse in nature; different questions ask about different and specific information about the image which leads to more diversity in learning and hence more transferable representations. For MM-IMDB the results vary, with COCO pretraining giving better performance with VisualBERT while CC-Small pretraining giving better results with ViLBERT. 
\begin{figure}[ht]
\centering
\includegraphics[width=\columnwidth]{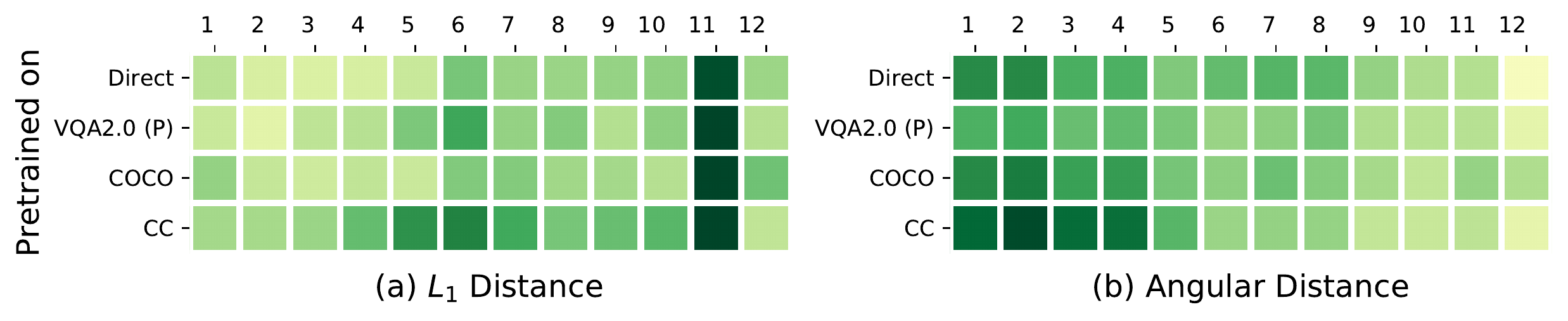}
\caption{$L_1$ distance (left) and angular distance (right) between pretrained and finetuned self-attention weights when using different pretrained datasets on downstream VQA-D dataset. Darker blocks represent higher distance. VQA-P representations are closer to the fine-tuned representations compared to other datasets and direct training.}
  \vspace{-8 pt}
\label{fig:vqa_vs_cc_visual_bert}
\end{figure}

To qualitatively analyze the transferability, we visualize the $L_1$ and angular distances between pretrained and finetuned self-attention layer weights for VisualBERT model in Fig.~\ref{fig:vqa_vs_cc_visual_bert}. The different weight matrices have been averaged per layer. We observe a larger drift in $L_1$ distances of weights for layers $5-10$ and in angular distances of weights for layers $1,2,3,4$ when finetuned from CC-small or COCO compared to finetuning from VQA-P. We also observe that lower layers ($1-6$) are more close in terms of $L_1$ while upper layers ($6-12$) are more close in terms of angular. These observations demonstrate that \textbf{representations trained from VQA-P} are more close to fine-tuned representations compared to other pretraining dataset and hence, \textbf{have greater transferability}. 

\subsection{Pretraining Scalability and a Promising Alternative}
\label{subsec:scalability}
In NLU, we have seen large gains on GLUE \cite{wang2018glue} and SuperGLUE \cite{wang2019superglue} just by increasing the amount of data the model was pretrained on \cite{liu2019roberta}. NLU pretraining depends on unlabelled text data ubiquitously present on the internet, while current methods for visio-linguistic pretraining are dependent on labelled data such as image-caption pairs. This hinders the push for large scale self-supervised pretraining as natural sources of aligned visio-linguistic data exist sparsely. Further, we noticed in Section~\ref{subsec:does_it_help} that even as we increase CC data for pretraining, the performance improvement is marginal. This is dissatisfying because we are unable to properly exploit the large scale of quality images available on the web due to inferior annotations. 
We propose a first step towards providing a solution to this scalability problem. We hypothesize that \emph{generated} data that is close to the domain of the downstream task can serve as a better pretraining dataset than  a natural but out-of-domain dataset. We empirically demonstrate this.
\begin{figure}[ht]
\centering
\includegraphics[width=\columnwidth]{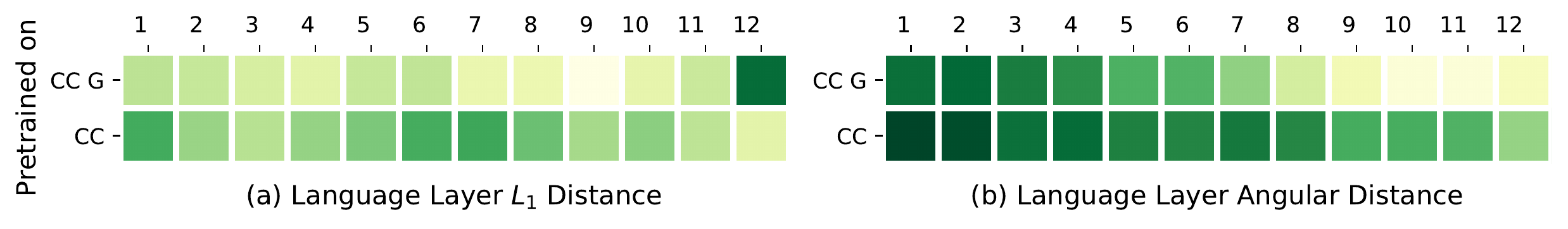}
\includegraphics[width=\columnwidth]{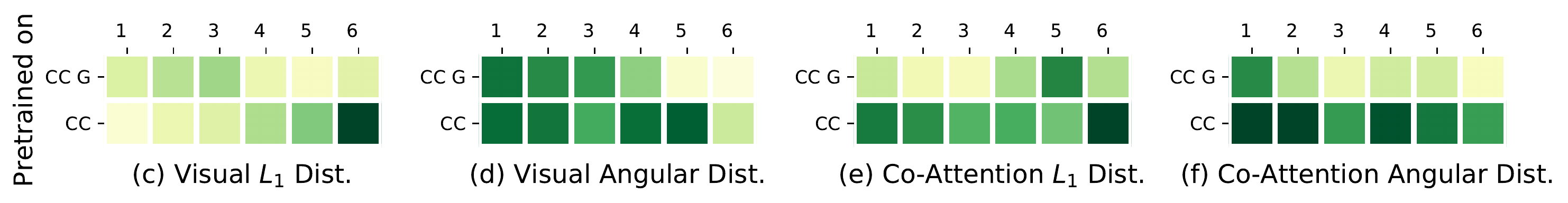}
\caption{$L_1$ distances and angular distances between CCG/CC pretrained and VQA-D finetuned self-attention weights' parameter spaces for the ViLBERT model averaged per layer. Darker blocks represent larger distance. Both $L_1$ and angular distances are \textbf{closer} for generated in-domain dataset (CCG) compared to out-of-domain dataset (CC).}
  \vspace{-8 pt}
\label{fig:cc_vs_cc_generated_vilbert}
\end{figure}
Specifically, we consider the downstream dataset VQA-D. We train a captioning model on a visually in-domain dataset (COCO). We then generate captions on an out-of-domain dataset (CC). Note that this generated dataset is now closer in domain to the downstream dataset (VQA-D) than CC. We find that it serves as a better pretraining dataset compared to using the original CC dataset with its associated “naturally” occurring textual data.

Fig.~\ref{fig:cc_vs_cc_generated}~(b,c) show performance on VQA-D after pretraining with VQA-P, COCO, CC and \textit{generated} CC (CCG). We see that pure in-domain pretraining always works best (VQA-P, COCO) and acts as an upper bound. \textit{Generated} CCG dataset performs better than out-of-domain CC dataset for both VisualBERT and ViLBERT; that is, it works irrespective of the model size, which is very encouraging. We also observe that increasing the size of the generated dataset further improves performance which sparks a possibility of scalability. While currently performance is worse than pretraining directly on COCO, there is promise that with a larger generated dataset, we may outperform it. In addition, perhaps training on a smaller COCO dataset may be sufficient to generate a sufficiently useful CCG; generating captions on images from Flickr (closer to COCO in visual domain than CC) may be even more effective. Exploring these directions is part of future work.

Following Section~\ref{subsec:transferability}, in Fig.~\ref{fig:cc_vs_cc_generated_vilbert} 
we observe that across all weights, the representations in parameter space are further apart for out-of-domain datasets. The pretrained weights from generated in-domain data, CCG, are more $L_1$ and angular close to their finetuned models. We provide more elaborate visualizations for both models in the supplementary.

\subsection{ViLBERT vs VisualBERT}
Throughout our experiments, we have observed that VisualBERT consistently outperforms ViLBERT under similar experimental settings even though ViLBERT uses an extra pretraining loss (masked image modelling) and has double the parameters. Note that as described earlier, we use fewer loss terms for both models than in the original paper because we did not find them to help performance in our settings. With that caveat, the trend in our experiments is quite consistent. We observe this behavior in 
(i) Fine-tuning where we see multiple $>$ 2\% accuracy drops in ViLBERT compared to VisualBERT (ii) Frozen base experiments where most ViLBERT metrics except VQA-D are lower than randomly initialized representations  (iii) CCG experiments where ViLBERT again has less accuracy than VisualBERT as seen in Fig.~\ref{fig:cc_vs_cc_generated}(c).

\subsection{Best pretrained model == Best downstream model?}
Often the choice of which pretrained model to use for downstream tasks is overlooked and with natural intuition one picks  the pretrained model with best validation pretraining loss. To analyze whether this choice works, we take three VisualBERT pretrained models with different validation pretraining losses (0.87, 0.95, 0.99) and evaluate them on VQA-D. We observe that the natural ordering of best pretrained to best fine-tuned is preserved in direct fine-tuning experiments though the error bars overlap: (69.91$\pm$0.02\%, 69.75$\pm$0.18\%, 69.43$\pm$0.18\%) while in the case of frozen base the order reverses: (48.64\%$\pm$0.19, 48.76\%$\pm$0.2, 49.57\%$\pm$0.18). This is surprising as this demonstrates that the best pretrained model may lead to best downstream model if finetuned but it may not be the most transferable and generalizable pretrained model otherwise.

\subsection{Does all this help in pushing the SoTA?}
Using the learnings of our systematic large-scale experiments and exhaustive analysis, we are able to achieve SoTA among published works for \textbf{SNLI-VE (test) - 77.57\%} and \textbf{MM-IMDB (test) - 68.04\% micro-F1}. Further, we achieve near SoTA results on \textbf{VQA 2.0 (test-dev) - 70.91\%} and \textbf{VizWiz (test) - 53.42\%}. All of these numbers are calculated using single models (no ensembles) without using extra data. For VQA 2.0 (i.e., VQA-D), these numbers are better than what were reported in the original VisualBERT \cite{li2019visualbert} and ViLBERT \cite{lu2019vilbert} papers. Further, the current state-of-the-art methods on VQA 2.0 uses a lot of tricks including more complex and extra image features (e.g. ResNet features), double pretraining, extra data and ensembles which does not allow for a fair comparison with our models.

\section{Related work}

\textbf{Self-Supervised PreTraining} In NLP, the general push has been towards increasing the size of the pretraining data and the model \cite{devlin2018bert, liu2019roberta} which has lead to better results on the downstream benchmarks \cite{wang2018glue, wang2019superglue}. Most of the pretraining schemes involve masking some part of the input and letting the model predict it. Masked Language Modeling, where a word of a sentence is masked with some probability, is the most common pretraining objective. Recent works have also shown that pretraining objectives matter but not as much as the amount of data and size of the model \cite{liu2019roberta}. Our paper empirically shows that in visio-linguistic pretraining increasing the data doesn't necessarily mean that it will improve the performance on the downstream task. In vision, \cite{doersch2015unsupervised} proposed a pretext task of predicting the relative location of image patches. This work spawned multiple works around predicting the ``jigsaw puzzle'' \cite{nathan2018improvements, noroozi2018boosting}. Other pretraining approaches include cleverly designed classification tasks such as predicting an image's orientation \cite{gidaris2018unsupervised}, classifying the label of image's cluster \cite{caron2018deep}, image inpainting \cite{pathak2016context}, image coloring \cite{zhang2016colorful} and motion segmentation prediction \cite{pathak2017learning}. 
\textbf{Visio-Linguistic Pretraining (V\&L)}
Contrary to earlier works in vision and language \cite{lu2016hierarchical, yu2018mattnet, lu2018neural, hu2017modeling, lee2018stacked, fukui2016multimodal, yu2019deep} which designed specialized architectures for different tasks, many recently introduced models for V\&L \cite{lu2019vilbert, li2019visualbert, tan2019lxmert, chen2019uniter, lu201912, alberti2019fusion, li2019unicoder, su2019vl, zhou2019unified} provide a common architecture that can be pretrained using self-supervised losses and adapted to many downstream V\&L tasks. Pretraining is performed on image captioning datasets like COCO Captions \cite{cococaption}, Conceptual Captions \cite{sharma2018conceptual}, SBU Captions \cite{yao2010i2t} etc. and then transferred to downstream task by fine-tuning the whole architecture end-to-end. Unlike self-supervised models in NLP like BERT\cite{devlin2018bert}, these models need to handle two types of data modalities: visual and textual. Recent works handle this in two ways, either use two transformer layer streams \cite{lu2019vilbert, lu201912, tan2019lxmert} for two modalities where the streams can interact later or project both modalities to a common space and then use a single stream \cite{li2019visualbert, chen2019uniter, li2019unicoder} on this combined projected feature space. Various pretraining self-supervised objectives have been used that include (i) Masked Image Caption Modeling (ii) Image Caption Matching (iii) Masked Object Feature Modeling etc. One major bottleneck with these setups is the need for \textit{labelled} data for pretraining which limits their scalability for large-scale self-supervised learning. In our work we show how to get around this bottleneck with synthetic data and also achieve $\sim$SOTA results leveraging some simple design choices during pretraining.

\section{Conclusion}

In this study, we perform an empirical analysis of visio-linguistic representations, by questioning various choices commonly made in the process of self-supervised pretraining for vision and language. We conduct experiments on a large scale using three different pretraining datasets and four downstream tasks to show that the source domain of the pretraining dataset matters and debunk the myth that visio-linguistic pretraining works out-of-box. 
We show that through the right pretraining choices we can achieve near state-of-the-art performance without using any extra data. We show how unlabelled data can be synthetically labelled to scale the pretraining for superior performance, taking a step towards relieving the bottleneck of labelled captioning data in visio-linguistic self-supervised pretraining. We hope that this work encourages future researchers to make the right choices for pretraining visio-linguistic architectures.

\section*{Acknowledgments}
We would like to thank Marcus Rohrbach for helpful discussions and feedback.
\bibliographystyle{splncs04}
\bibliography{egbib}

\newpage
\appendix 

\counterwithin{figure}{section}
\counterwithin{table}{section}
\counterwithin{equation}{section}

\begin{center}
\Large 
\textbf{Are we pretraining it right? \\ Digging deeper into visio-linguistic pretraining}\\(Supplementary Material)
\par
\end{center}

We provide additional details about (\ref{supp:exp_details}) experimental setup; (\ref{supp:pretraining}) pretraining model performance; (\ref{supp:distances}) $L_1$ and angular distances; (\ref{supp:ccg}) effect of synthetic data on downstream tasks and (\ref{supp:cc_vs_coco_comparison}) domain comparison between CC and COCO pretraining datasets.

\section{Details of Experimental Setup}

In Table~\ref{tab:hyperparam}, we list out the different hyperparameter settings for our pretraining experiments. The exact reproduction details of downstream fine-tuning experiments will be made available with code release.

\label{supp:exp_details}

\begin{table}[ht]
\centering
\scriptsize
\caption{Table with hyperparameters for different pretraining setups.}
\smallskip
\setlength\tabcolsep{3 pt}
  \begin{tabular}{ c | c | c | c | c | c | c }
  \toprule
    \multicolumn{1}{c|}{\bf Pretrained On} &
    \multicolumn{2}{c|}{\bf COCO} &  \multicolumn{2}{c|}{\bf VQA-P} &  \multicolumn{2}{c}{\bf CC}\\
      
      \cmidrule(lr){1-1} 
      \cmidrule(lr){2-3}
      \cmidrule(lr){4-5}
      \cmidrule(lr){6-7}
    \bf HyperParam & \bf VisualBERT &
    \bf ViLBERT & \bf VisualBERT &
    \bf ViLBERT & \bf VisualBERT &
    \bf ViLBERT \\
    \midrule
    Number of Layers & 12 & T-12, V-6 & 12 & T-12, V-6 & 12 & T-12, V-6 \\
    Batch Size & 896 & 896 & 896 & 1024 & 896 & 1024  \\
    Max Updates & 88,000 & 88,000 & 88,000 & 88,000 & 88,000 & 88,000 \\
    Peak Learning Rate & 5e-5 & 5e-5  & 5e-5  & 5e-5  &  5e-5 & 5e-5 \\
    Hidden Size & 768 & T-768, V-1024 & 768 & T-768, V-1024  & 768 & T-768, V-1024\\
    FFN inner Hidden Size & 3072 & T-3072, V-1024 & 3072 & T-3072, V-1024 & 3072 & T-3072, V-1024 \\
    Dropout & 0.1 & 0.1  & 0.1 & 0.1 & 0.1 & 0.1\\
    Attention Dropout & 0.1 & 0.1  & 0.1 & 0.1 & 0.1 & 0.1 \\
    Warmup Steps & 2000 & 2000& 2000 & 2000& 2000 & 2000 \\
    Learning Rate Decay & Cosine & Cosine  & Cosine & Cosine & Cosine & Cosine\\
    Adam $\epsilon$ & 1e-8 & 1e-8  & 1e-8 & 1e-8 & 1e-8 & 1e-8\\
    Adam $\beta_1$ & 0.9 & 0.9 & 0.9 & 0.9 & 0.9 & 0.9\\
    Adam $\beta_2$ & 0.98 & 0.98 & 0.98 & 0.98 & 0.98 & 0.98\\
    Gradient Clipping & 0 & 0 & 0 & 0 & 0 & 0 \\
  \bottomrule
  \end{tabular}
  \vspace{-1mm}
  \label{tab:hyperparam}
\end{table}

\section{Pretraining}
\label{supp:pretraining}

\textbf{Performance.} In this section, we see how well the models perform on pretraining objectives for different pretraining datasets. Table~\ref{tab:pretrain} (left) shows the best validation losses that were achieved after finetuning hyperparameters for each of these datasets. It is important to note here that models sees equal number of training samples for each of the pretraining datasets and was trained for the same number of training iterations (88K). We observe that both models achieve $\sim$ lowest MLM loss using COCO. We hypothesize that this lower pretraining loss points to less noisy and superior quality annotations. ViLBERT achieves lowest MIM\footnote{Masked Image Modelling loss. ViLBERT's MMM loss can be broken down into MLM and MIM.} loss on CC-Small though only slightly better than COCO and VQA-P, both of which have a similar loss due to same images. 
\textbf{Pretraining Dataset Size.} Table~\ref{tab:pretrain} (right) shows the effect on pretraining objectives as we increase the dataset size. We observe that increasing dataset size improves validation performance consistently. Here we increase dataset size for CC in subsets of 10\%, 50\% and 100\%. Note that the number of epochs in these set of experiments is kept constant (30 Epochs) so as to not overfit smaller size datasets. We see the difference in CC 10\% on left and right due to difference in the number of iterations of training (88K on left vs 12K on right), indicating longer pretraining is helpful.

\begin{table}[t]
\footnotesize
\centering
  \caption{\textbf{Pretraining validation losses} of VisualBERT and ViLBERT. MLM points to masked language modelling object and MIM points masked image modelling objective. \textbf{(Left)} Losses on different same size pretraining datasets when pretrained for the same number of iterations. COCO and CC achieve the lowest and highest losses respectively. \textbf{(Right)} Losses on different sized subsets of CC. Adding more data consistently improves validation performance when trained for the same number of epochs. The validation set for all CC datasets are the same.}
\smallskip
\setlength\tabcolsep{10 pt}
\resizebox{\columnwidth}{!}{
  \begin{tabular}{c | c c c | c c c }
  \toprule
         & \multicolumn{3}{c}{Same Iterations} & \multicolumn{3}{c}{Same Epochs} \\
    \cmidrule(r){2-4}
    \cmidrule(r){5-7}
     Model & COCO  & VQA 2.0 (P) & CC 10\%  & CC 10\%  & CC 50\%  & CC 100\% \\
                
    \\      \midrule
    VisualBERT(MLM) & 0.87  & 0.91  & 1.97 & 2.22 & 2.00 & 1.96 \\
    \midrule
    ViLBERT(MLM) & 0.98 & 1.01 & 2.03 & 2.33 & 2.11 & 1.93 \\
    ViLBERT(MIM) & 0.19 & 0.20 & 0.17 & 0.35 & 0.22 & 0.17 \\
  \bottomrule
  \end{tabular}}
  \label{tab:pretrain}
  \vspace{-8 pt}
\end{table}

\section{$L_1$ and Angular Distances}
\label{supp:distances}

 In this section we provide more details about the $L_1$ and angular distances between the pretrained and finetuned models for various experiments in Section~3.2 
and Section~3.4 .

\begin{figure}[ht]
\centering
\includegraphics[width=\columnwidth]{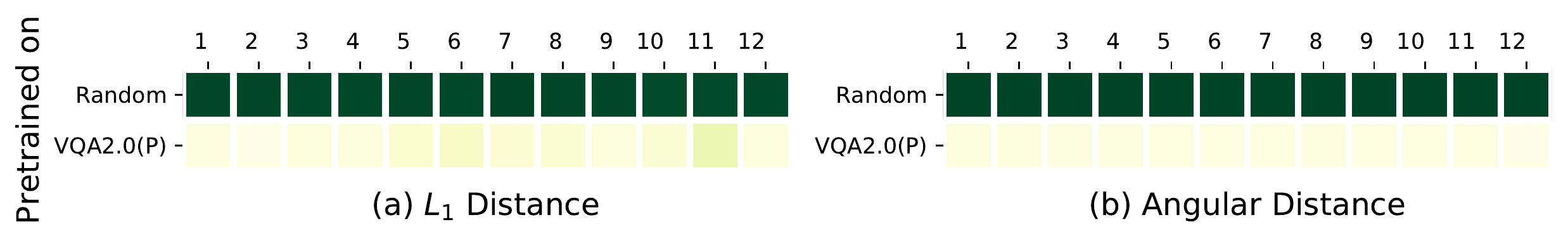}
\caption{$L_1$ distances and angular distances for randomly initialized and VQA-P pretrained model and their finetuned self-attention weights' respectively for the VisualBERT model averaged per layer. Darker blocks represent larger distances. Both $L_1$ and angular distances are much \textbf{closer} for VQA-P compared to randomly initialized model.}
  \vspace{-8 pt}
\label{fig:supp_random_vs_vqa_visual_bert}
\end{figure}

\textbf{Random vs Pretrained.} In Fig.~\ref{fig:supp_random_vs_vqa_visual_bert}, we show how the finetuned representations change if we start from a randomly initialized model compare to a pretrained model(here VQA-P). Note that a randomly initialized model is different from direct training. In direct training, the models are initialized with BERT weights as mentioned in Sec 2.5. However, with randomly initialized finetuning, the initialized weights are from a normal distribution with mean and standard deviation of 0.0 and 0.02 respectively. We observe that pretrained self-attention weights are much closer in both $L_1$ and angular distance compared to randomly initialized ones.

\begin{figure}[ht]
\centering
\includegraphics[width=\columnwidth]{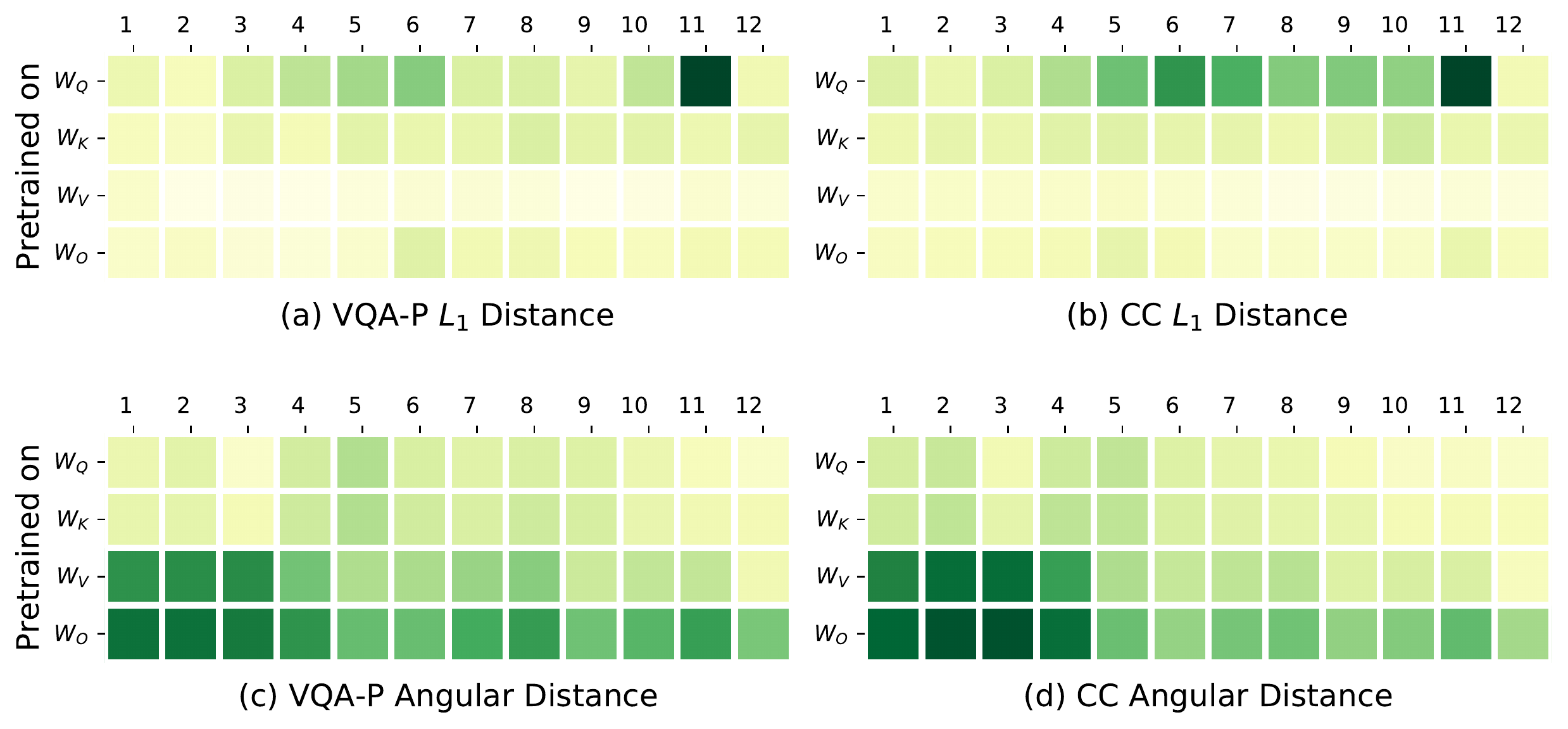}
\caption{Breakdown by individual self-attention weights showing $L_1$ and angular distances between pretrained and finetuned models for VQA-P and CC VisualBERT models. For CC, $W_Q$ $L_1$ distances and $W_V, W_O$ angular distances are further apart compared to VQA-P. Darker blocks represent more distance between the representations \textit{i.e.} more further apart.} 
  \vspace{-8 pt}
\label{fig:supp_vqa_vs_cc_visual_bert}
\end{figure}

\textbf{VisualBERT.} In Fig.~\ref{fig:supp_vqa_vs_cc_visual_bert}
, we show more details about how self-attention weights change from pretraining to finetuning. We show the breakdown for the self-attention weights $W_Q$, $W_K$, $W_V$, $W_O$ for all 12 transformer blocks. From Fig.~4 in the main paper, we pick VQA-P and CC pretrained models and show how they change after finetuning on VQA-D. We observe that for CC pretrained, the $W_Q$ weights are more $L_1$ \textbf{distant} compared to VQA-P. For angular distances, $W_V$ and $W_O$ weight matrices are more \textbf{closer} for VQA-P compared to CC.

\begin{figure}[ht]
\centering
\includegraphics[width=\columnwidth]{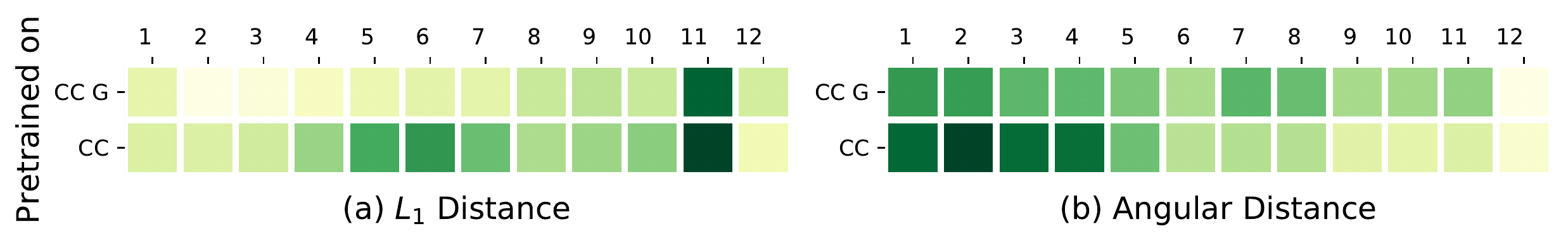}
\caption{$L_1$ and angular distances for CC and CCG for VisualBERT model. Both $L_1$ and angular distances are \textbf{closer} for CCG compared to CC pretrained similar to what we observed in Fig.~4 for ViLBERT. Darker blocks represent more distance between the representations \textit{i.e.} more further apart.}
  \vspace{-8 pt}
\label{fig:supp_cc_vs_ccg_visual_bert}
\end{figure}

In Sec.~3.4 and
Fig.~5
in the main paper, we showed the $L_1$ and angular distances for CC and generated CC for ViLBERT model. In Fig.~\ref{fig:supp_cc_vs_ccg_visual_bert}, we show the same for VisualBERT model and observe a similar pattern. The pretrained weights from generated in-domain data, CCG, are more $L_1$ and angular close to their finetuned models agnostic of model architecture.

\section{Synthetic data pretraining}
\label{supp:ccg}

In this section, we show the performance of synthetic data on more downstream tasks. We pretrain on the same \textit{generated} CC (CCG) dataset and finetune on Vizwiz and SNLI-VE. We observe that CCG gives a better downstream performance on Vizwiz for VisualBERT and close to CC for ViLBERT. For SNLI-VE, we see comparable but not better performance for CCG which can be attributed to the fact that generated captions for CCG are closer in domain to COCO/VQA-D but not SNLI-VE. As CCG pretraining performs competitively or better than CC pretraining for all of the tasks, it further solidifies our claim in Sec~3.4. 
in the main paper, that synthetic data can help scale up visio-linguistic pretraining.

\begin{table}[ht]
\footnotesize
\centering
  \caption{Results of CCG pretraining on Vizwiz and SNLI-VE. CCG pretraining performs competitively or better than CC pretraining for VizWiz and SNLI-VE as well.}
\smallskip
\setlength\tabcolsep{10 pt}
  \begin{tabular}{c | c | c  c  c  }
  \toprule
      & & VQA-D & Vizwiz & SNLI-VE \\
      \cmidrule(lr){3-3}
      \cmidrule(lr){4-4}
      \cmidrule(lr){5-5}
     Pretrained on & Model  & acc & acc & acc  \\
      \midrule
    CC & VisualBERT & 68.91  & 52.37 & \bf 77.33 \\
    CCG & VisualBERT & \bf 68.97  & \bf 53.04  & 77.05 \\
  \midrule
    CC & ViLBERT & 68.38  & \bf 52.01  & \bf 75.67 \\
    CCG & ViLBERT & \bf 68.84  & 51.71  & 75.53 \\
  \bottomrule
  \end{tabular}
  \label{tab:ccg}
\end{table}

\section{Conceptual Captions and COCO Domain Comparison}
\label{supp:cc_vs_coco_comparison}
\begin{figure}[ht]
\centering
\includegraphics[width=\columnwidth]{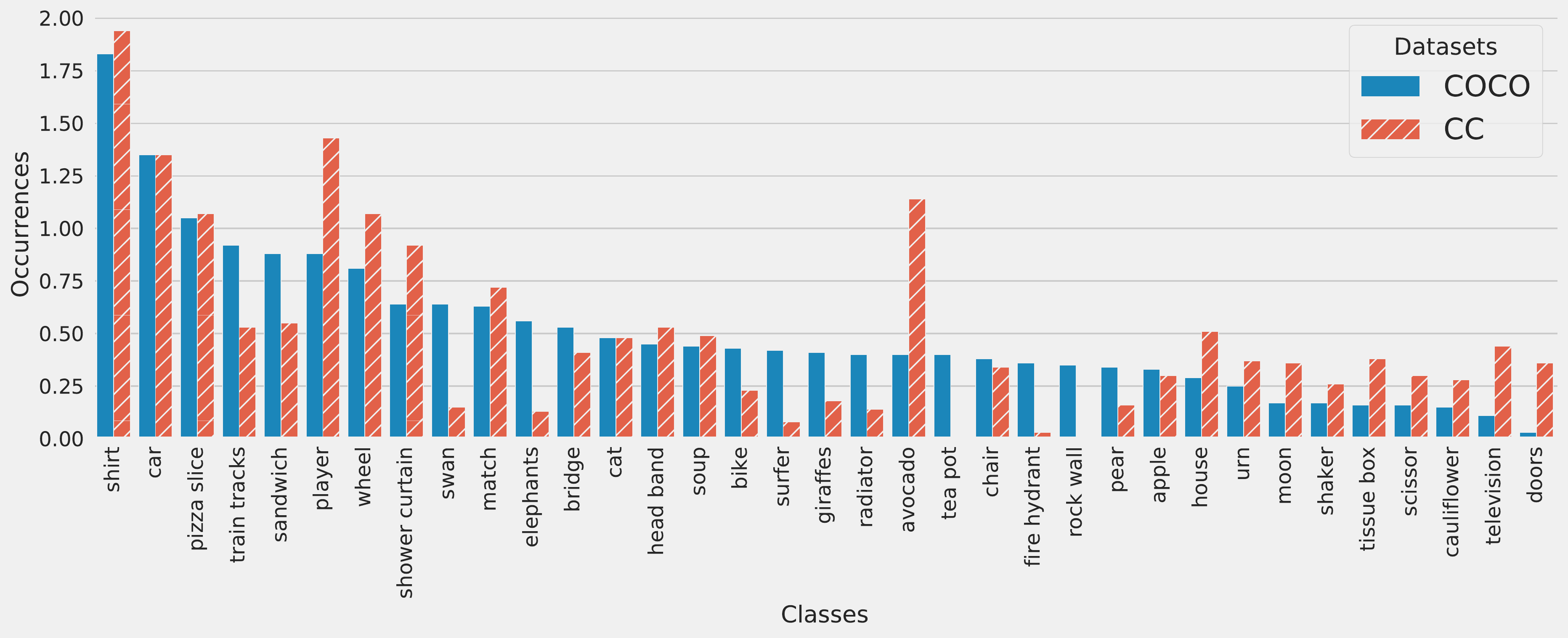}
\caption{Average number of instances of \textbf{top occurring Visual Genome categories} in COCO and CC in random 100 images from validation sets of both COCO and Conceptual Captions. The visual domain of both datasets are \textbf{contrasting} as can be seen from variation in occurrence of different categories.}
\end{figure}
In this section, we compare the visual domains of CC and COCO to show that they are different. For this purpose, we compare the object instances between both datasets. Since CC doesn't have any object bounding box annotations, we rely on an object detector trained on VisualGenome dataset which contains 1600 object classes to extract object bounding box proposals. We extract 100 object proposals from each image to ensure maximum coverage. From these 100 object proposals, we ignore background class objects. COCO on average contains 44.46 objects per image while CC on average contains 31.95 objects per image which depict that COCO generally contains a vast and diverse set of images compared to CC which contains Wikipedia-style images that are frequently focused only on few objects. Further, we compare the top occurring Visual Genome categories in COCO and CC in Fig.~\ref{supp:cc_vs_coco_comparison} by selecting random 100 images from validation sets of both COCO and CC. We observe a stark difference in the distribution which shows that both datasets are indeed contrasting.

\end{document}